# Accurate Prediction of Temperature Indicators in Eastern China Using a Multi-Scale CNN-LSTM-Attention model

**Jiajiang Shen[1,4], Weiyan Wu[2,5], Qianyu Xu[3,6]**

[1] School of Artificial Intelligence and Computer Science, Nantong University, Jiangsu, China
[2] Fordham University, New York, USA
[3] Medill School of Journalism, Northwestern University, USA

[4] 1844006250@qq.com
[5] wwu51@fordham.edu
[6] qianyuxu2017@u.northwestern.edu

**Abstract.** In recent years, the importance of accurate weather forecasting has become increasingly prominent due to the impacts of global climate change and the rapid development of data science. Traditional forecasting methods often struggle to handle the complexity and nonlinearity inherent in climate data. To address these challenges, we propose a weather prediction model based on a multi-scale convolutional CNN-LSTM-Attention architecture, specifically designed for time series forecasting of temperature data in China. The model integrates Convolutional Neural Networks (CNN), Long Short-Term Memory (LSTM) networks, and attention mechanisms to leverage the strengths of spatial feature extraction, temporal sequence modeling, and the ability to focus on important features. The development process of the model includes data collection, preprocessing, feature extraction, and model building. Experimental results show that the model performs excellently in predicting temperature trends with high accuracy. The final computed results indicate that the Mean Squared Error (MSE) is 1.978295 and the Root Mean Squared Error (RMSE) is 0.8106562. This work marks a significant advancement in applying deep learning techniques to meteorological data, offering a valuable tool for improving weather forecasting accuracy and providing essential support for decision-making in areas such as urban planning, agriculture, and energy management.

## 1. Introduction

With the rapid advancements in global climate change and data science, time series forecasting has become an increasingly pivotal tool in fields such as meteorology, environmental science, and urban planning. As one of the key factors influencing residents' quality of life and urban planning, accurate forecasting of weather temperature is of great significance to government decision-making, agricultural production, energy management, urban planning and infrastructure construction, public health management, and many other areas. However, traditional meteorological forecasting methods are often constrained by the complexity of data and the inherent limitations of the models themselves, making it difficult to conduct multi-scale predictions amidst the challenges of missing early meteorological data, the impacts of climate change, and the high-dimensional nonlinear characteristics of meteorological data. Consequently, these methods often fail to fully capture the dynamics and nonlinear features of climate change.

In recent years, the emergence of deep learning technologies has provided new solutions for time series forecasting. Techniques like Back Propagation Neural Networks (BPNN) [1], Recurrent Neural Networks (RNN) [2], and Random Forest algorithms [3] have been applied to weather time series forecasting, significantly advancing the automation of meteorological predictions. Notably, the combination of Convolutional Neural Networks (CNN) and Long Short-Term Memory networks (LSTM), known as the CNN-LSTM model, has demonstrated remarkable performance across multiple domains due to its powerful feature extraction capabilities and its ability to capture long-term dependencies in time series data.

Building on the CNN-LSTM framework, this study introduces a multi-scale CNN-LSTM-Attention model for weather forecasting. The attention mechanism enhances the model's ability to focus on the most relevant temporal features, improving its capacity to capture dynamic relationships in meteorological time series data. By assigning different weights to time steps, the model prioritizes important temporal information, boosting forecasting accuracy. The diverse climate of China makes it an ideal region for this research, which addresses challenges such as missing data and high-dimensional processing. This study contributes to both academic and practical advancements, providing valuable insights for weather forecasting in regions with similar climates, and sets the stage for future developments in meteorological prediction.

## 2. Related Work

Recent advancements in weather prediction and time series forecasting, driven by machine learning, particularly deep learning, have significantly improved the accuracy and efficiency of forecasting models. These techniques have enhanced the precision of temperature and weather pattern predictions while automating the process. The CNN-LSTM method, widely used in traffic congestion prediction and aircraft flight trajectory analysis, has proven effective in capturing complex temporal patterns. Below is an overview of key developments in this domain, highlighting the integration of deep learning approaches in diverse forecasting applications.

N. Ranjan et al [4]. introduced the Seoul Transportation Operation and Information Service (TOPIS) and a hybrid model combining CNN, LSTM, and Transpose CNN for predicting network-wide congestion levels. This model effectively extracts spatial and temporal information from images, demonstrating improved precision, recall, and accuracy across various prediction horizons. The PredNet model achieved a performance improvement of 2% to 12% in road-wise predictions.

YU Zhang, Qingxia He, and Yishi Zen developed a CNN-LSTM model to predict global annual average temperature over the next 20 years, using global temperature data from 1880 to 2022 [5]. Their results highlight the model's strengths, with CNN reducing data dimensionality and LSTM capturing long-term dependencies, enhancing prediction accuracy for large-scale temperature data. Similarly, CJ Huang and PH Kuo combined CNN and LSTM to predict PM2.5 concentration, demonstrating that their CNN-LSTM model (APNet) outperforms other machine learning methods in accuracy, using historical data such as rainfall, wind speed, and PM2.5 concentrations for forecasting [6].

Bao Wang, Shichao Liu, and Bin Wang employed CNN-LSTM for short-term storm surge prediction. This study used CNN and LSTM both individually and in combination for multi-step ahead short-term storm surge level prediction, utilizing observed SL and wind data. A case study with 11 years of hourly SL and wind data from Xiuying Station, Hainan Province, China, was conducted to validate the models. The results show that CNN and LSTM outperform Support Vector Regression (SVR) and Multilayer Perceptron (MLP)[7], with the combined models improving accuracy by 4% to 6%. During two severe typhoons, the model's accuracy increased by over 10% across all forecasting steps.

Wang, B., Liu, S., and Wang proposed a novel 4D trajectory prediction hybrid architecture based on deep learning, combining CNN and LSTM[8]. The model employs 1D convolution to extract spatial features and LSTM to capture temporal features, resulting in high-precision 4D trajectory prediction.

Experimental results show that the CNN-LSTM hybrid model outperforms single models, reducing prediction error by an average of 21.62% compared to the LSTM model and 52.45% compared to the BP model.

**3. Predicting Temperature Model**

The dataset encompasses various weather indicators recorded every hour from 00:00 to 24:00 in the eastern region of China from 2001 to 2020, including weather conditions, dew point temperature, fog, hail, heat index, relative humidity, precipitation, barometric pressure, rainfall, snowfall, temperature, thunderstorms, tornadoes, visibility, wind direction (mentioned twice accidentally, likely a typo), surface wind speed, wind chill factor, and wind speed, totaling 182,991 records. We also analyzed the distribution of the data through charting, such as the distribution of weather conditions, wind directions, and temperatures, as shown in Figures 1. These visualizations aided in understanding the fundamental characteristics and potential patterns within the dataset.

*3.1. Data Cleaning and Resampling*

During data processing, temporal feature extraction was performed by converting date and time information into sub-features such as year, month, and day. Temperature data was standardized using MinMaxScaler, scaling it to the range of -1 to 1 to eliminate dimensional discrepancies. Missing values for continuous variables, like temperature and humidity, were imputed using the mean. Low-quality or incorrect data segments were manually deleted to ensure dataset reliability. A sliding window approach was used to construct input-output sequences, where each input sequence included 30 days of data to predict the temperature for the 31st day, preserving temporal trends.

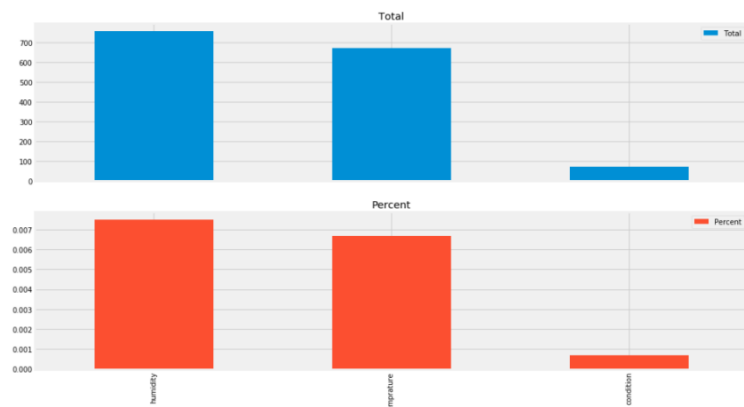

**Fig.1** Visualization of Data Distribution

*3.2. Model Architecture*

In weather forecasting, traditional CNN and LSTM models have limitations. CNNs struggle with temporal dependencies in time series data, while LSTMs have difficulty extracting spatial features and are computationally expensive for long sequences. To address these challenges, we propose a CNN-LSTM-Attention model, which combines CNN, LSTM, and Attention mechanisms to process both spatial and temporal features effectively.

    CNN Layer: Extracts spatial features from weather data, capturing local patterns like temperature distribution and wind speed. The multi-scale convolution design helps capture spatial information at different levels.

    LSTM Layer: Captures long-term temporal dependencies in the data, learning how past weather events influence future ones.

Attention Mechanism: Focuses the model's attention on the most important time steps when making predictions, improving accuracy by assigning higher weights to relevant data. This mechanism also enhances interpretability by showing which parts of the data the model prioritizes.

By combining CNN, LSTM, and Attention, the model effectively captures both spatial and temporal dependencies in weather data, improving prediction accuracy. It also handles multi-dimensional data, integrating multiple meteorological variables for more comprehensive forecasts. This approach significantly enhances performance compared to traditional models that process only a single data type.

**Table1 .** CNN-LSTM-Attention Model Architecture

| Layer | Out Shape |
|---|---|
| Input_layer | (None,30 ,1) |
| Conv1D | (None,29, 256) |
| Conv1D | (None, 28, 128) |
| MaxPooling1D | (None, 14, 128) |
| Flatten | (None, 1792) |
| RepeatVector | (None, 30, 1792) |
| LSTM | (None, 30, 100) |
| Dropout (rate=0.3) | (None, 30, 100) |
| LSTM | (None, 30, 100) |
| Dropout (rate=0.3) | (None, 30, 100) |
| LSTM | (None, 30, 100) |
| Bidirectional LSTM | (None, 30, 256) |
| Self-Attention |  |
| Dense (units=100) | (None, 30, 100) |
| Dense (units=1) | (None, 1) |

The proposed model combines multi-layer convolutions and LSTMs, structured as a CNN-LSTM-Attention network using Keras's Sequential API. The convolutional layers, with 256 and 128 filters of size 2, extract local features like seasonal trends, followed by ReLU activations for nonlinearity. MaxPooling layers reduce feature map dimensions to enhance model efficiency. To capture long-term dependencies, we employ multi-layer bidirectional LSTMs with 100 units, interspersed with Dropout layers to prevent overfitting. The bidirectional LSTMs enhance the model's ability to capture both forward and backward dependencies, improving generalization. Additionally, an Attention Mechanism is incorporated, dynamically weighting different time steps to focus on the most relevant data, thus improving prediction accuracy. The final output is mapped to a single neuron via a fully connected layer, enabling accurate daily temperature predictions. This architecture is designed to optimize both feature extraction and temporal dependency modeling for more precise forecasting.

### 3.3. Training Strategy

During model compilation and training, we selected Mean Squared Error (MSE) as the loss function and employed the NAdam optimizer for parameter optimization. Compared to the traditional Adam optimizer, NAdam combines the momentum method with Nesterov Accelerated Gradient (NAG), which allows for more flexible adjustments in the direction of parameter updates. This feature is particularly beneficial when dealing with complex nonlinear features, as it demonstrates faster convergence and greater stability. Additionally, the NAdam optimizer is more sensitive to learning rate adjustments, helping maintain efficient optimization throughout the model training process.

We set an initial learning rate (INIT_LR) for the NAdam optimizer and employed a decay strategy (decay = INIT_LR/EPOCHS) to finely adjust the learning rate throughout the training. During the training process, we monitored the loss function changes and implemented Early Stopping to prevent overfitting, effectively controlling the training process. Finally, predictions on the test set were compared with actual values, enabling a direct assessment of the model's predictive ability, and MSE was calculated as the evaluation metric.

## 4. Experiments

As shown in Figure 3, the predicted red curve closely aligns with the actual blue curve across most time steps, demonstrating high consistency in overall trends and minimal errors between predicted and actual values. These errors remain stable throughout the prediction period, with no systematic deviations, such as persistent over- or under-predictions. The model effectively captures real-time or near real-time data changes without significant time delays. Fluctuations and inflection points in the predicted data align well with similar patterns in the actual data, showcasing the model's adaptability and generalization to complex data patterns.

The integration of the Attention Mechanism allows the model to focus on time steps with the greatest influence on predictions. By assigning dynamic weights, the mechanism identifies and prioritizes key features, enhancing prediction accuracy, especially in cases with nonlinear or intricate patterns. This capability significantly improves the model's ability to detect subtle changes in the data.

While slight deviations occur near extreme peaks and valleys, likely due to the model's sensitivity limitations or unaccounted complexity in the training data, the overall performance demonstrates its strength in capturing primary trends. This confirms the effectiveness of combining CNN, LSTM, and Attention Mechanism for modeling complex time series data.

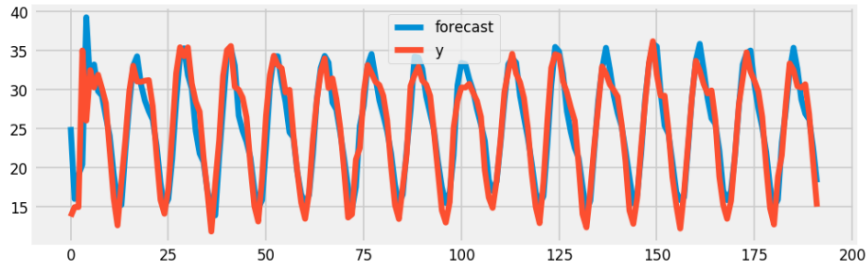

**Fig.3** Test Value and Predictive Value

The model's performance on the test set was evaluated and compared to baseline models, demonstrating the superiority of the hybrid architecture in handling time series data. Specific evaluation metrics included Mean Squared Error, Root Mean Squared Error, providing standard measures of prediction error. The formulas were as follows:

$$\text{MSE} = \frac{1}{n}\sum_{i=1}^{n}(y_i - \hat{y}_i)^2$$

（2）

$$\text{RMSE} = \sqrt{\text{MSE}}$$

（3）

These metrics measured the prediction error of the model. In this study, the calculated Mean Squared Error was 1.978295. RMSE was 0.8106562. Given that the temperature data in this experiment ranged from 0 to 100, the relatively small MSE value indicated that the model's predictions were close to the actual values.

## 5. Conclusions

Through this study, it was found that the use of the multi-scale convolutional CNN-LSTM-Attention model can effectively predict the time series of temperature data in the eastern region of China. Specifically, the CNN layer extracts local features from the temperature data, while the LSTM layer captures long-term dependencies in the time series, enabling the model to predict future temperatures more accurately. To further improve the model's performance, we incorporated an attention mechanism. The role of this mechanism is to help the model focus on the most important parts of the input sequence, particularly in capturing long-term dependencies and dynamic changes. By assigning different weights to each time step, the attention mechanism allows the model to prioritize the segments of the sequence that have a greater impact on the prediction, thereby improving the model's performance in complex meteorological data. This approach enables the multi-scale convolutional CNN-LSTM-Attention model to more accurately capture seasonal and trend variations in temperature.

As a result, the multi-scale convolutional CNN-LSTM-Attention model performs well in capturing trends and seasonal characteristics of temperature changes, particularly in identifying seasonal and trend shifts. The Mean Squared Error (MSE) of the final model on the test set is 1.978295, which is considered an ideal level.

This study demonstrates the potential of the multi-scale convolutional CNN-LSTM-Attention model in meteorological time series prediction, filling some gaps in the field of complex meteorological data prediction, especially in the application of combining convolutional neural networks, long short-term memory networks, and the attention mechanism. The findings provide valuable reference and inspiration for other researchers exploring the application of deep learning in meteorological prediction, and offer practical experience for research on combining different neural network models for time series forecasting. By introducing the attention mechanism, the model can more intelligently select and process the most relevant information, leading to improved prediction accuracy and stability.

However, the current research mainly focuses on single temperature prediction and has not yet fully incorporated the prediction of other meteorological factors. Future research should include more meteorological parameters in model training, focus on multivariate time series forecasting, and introduce more complex model architectures to enhance the model's generalization ability and prediction accuracy.